\documentclass[preprint,11pt]{elsarticle}

\usepackage{amssymb}

\journal{Neural Networks}

\usepackage{graphicx}
\usepackage{amsmath}
\usepackage{multirow}
\usepackage{subfig}
\usepackage{color,soul}
\usepackage{geometry}
\begin{document}

\begin{frontmatter}



\title{\textbf{Representation Learning using Event-based STDP}}


\author[l1]{Amirhossein Tavanaei}
\author[l2]{Timoth\'{e}e Masquelier}
\author[l1]{Anthony Maida}
\address[l1]{
The School of Computing and Informatics\\
University of Louisiana at Lafayette,
Lafayette, LA 70504, USA\\
Email: \{tavanaei,maida\}@louisiana.edu}
\address[l2]{CERCO UMR 5549,\\
CNRS-Universit\'{e} de Toulouse 3,
F-31300, France\\
Email: timothee.masquelier@cnrs.fr}

\begin{abstract}
Although representation learning methods developed within the framework of traditional neural networks are relatively mature, developing a spiking representation model remains a challenging problem. This paper proposes an event-based method to train a feedforward spiking neural network (SNN) layer for extracting visual features. The method introduces a novel spike-timing-dependent plasticity (STDP) learning rule and a threshold adjustment rule both derived from a vector quantization-like objective function subject to a sparsity constraint. The STDP rule is obtained by the gradient of a vector quantization criterion that is converted to spike-based, spatio-temporally local update rules in a spiking network of leaky, integrate-and-fire (LIF) neurons. Independence and sparsity of the model are achieved by the threshold adjustment rule and by a softmax function implementing inhibition in the representation layer consisting of WTA-thresholded spiking neurons. Together, these mechanisms implement a form of spike-based, competitive learning. Two sets of experiments are performed on the MNIST and natural image datasets. The results demonstrate a sparse spiking visual representation model with low reconstruction loss comparable with state-of-the-art visual coding approaches, yet our rule is local in both time and space, thus biologically plausible and hardware friendly.
\end{abstract}


\end{frontmatter}

\section{Introduction}
Unsupervised learning approaches using
neural networks have frequently been used to extract features from visual inputs~\cite{bhand2011unsupervised,lee2008sparse}.
Single layer networks using distributed representations or autoencoder networks~\cite{coates2011analysis,bengio2013representation} have offered effective representation platforms. 
However, the robust, high level, and efficient representation that is obtained by networks in the brain is still not fully understood~\cite{quiroga2005invariant,logothetis1996visual,riesenhuber2002neural,young1992sparse,landi2017two,wandell1995foundations,fregnac2016visual}. Understanding the brain's functionality in representation learning can be accomplished by studying spike activity~\cite{self2016effects} and bio-inspired spiking neural networks (SNNs)~\cite{maass1997networks,izhikevich2004model,ghosh2009spiking}.
SNNs provide a biologically plausible architecture, high computational power, and an efficient neural implementation~\cite{maass2015spike,maass1996computational,neil2016learning}.
The main challenge is to develop a spiking representation learning model that encodes input spike trains to uncorrelated, sparse, output spike trains using spatio-temporally local learning rules. 

In this study, we seek to develop representation learning in a network of spiking neurons to address this challenge. Our contribution determines novel spatio-temporally local learning rules embedded in a single layer SNN to code independent features of visual stimuli received as spike trains. Synaptic weights in the proposed model are adjusted based on a novel spike-timing-dependent plasticity (STDP) rule which achieves spatio-temporal locality. 

Nonlinear Hebbian learning has played a key role in the development of a unified unsupervised learning approach to represent receptive fields~\cite{brito2016nonlinear}.  
F\"{o}ldi\'{a}k~\cite{foldiak1990forming},
influenced by Barlow \cite{Barlow1989a}, 
was one of the early designers of sparse, weakly distributed representations having
low redundancy.
F\"{o}ldi\'{a}k's model introduced a set of three learning rules (Hebbian, anti-Hebbian, and homeostatic) to work in concert
to achieve these representations.
Zylberberg et al.~\cite{zylberberg2011sparse}
showed that F\"{o}ldi\'{a}k's plasticity rules, in a spiking platform, could be derived
from the constraints of reconstructive accuracy, sparsity, and decorrelation.
Furthermore,
the acquired receptive fields of the representation cells in their model (named SAILnet) qualitatively matched those
in primate visual cortex. The representation kernels determining the synaptic weight sets have been successfully utilized by our recent study~\cite{tavanaei2017multi} for a spiking convolutional neural network to extract primary visual features of the MNIST dataset. Additionally, the learning rules only used information which was locally
available at the relevant synapse.
Although SAILnet utilized spiking neurons in the representation layer and the plasticity rules were spatially local, the learning rules
were not temporally local.
The SAILnet plasticity rules use spike counts accumulated over the duration of a stimulus presentation interval. Since the SAILnet rules do not use spike times, the question of
training the spiking representation network using a spatio-temporally local, spike-based approach like spike-timing-dependent plasticity (STDP)~\cite{markram2012spike}, which needs neural spike times, remains unresolved.
Later work, \cite{king2013inhibitory}, extends~\cite{zylberberg2011sparse} to use both excitatory and inhibitory neurons (obeying
Dale's law), but the learning rules still use temporal windows of varying duration
to estimate spike rates, rather than the timing of spike events. Our work seeks to develop a learning rule which matches this performance but remains local in both time and space.

In another line of research based on cost functions, Olshausen and Field \cite{olshausen1996emergence}
and Bell and Sejnowski \cite{bell1997independent} showed that the constraints of
reconstructive fidelity and sparseness, when applied to natural images, could 
account for many of the qualitative receptive field (RF) properties of
primary visual cortex (area 17, V1).
These works were agnostic about the possible
learning mechanisms used in visual cortex to achieve these representations. Following~\cite{olshausen1996emergence}, Rehn and Sommer~\cite{rehn2007network} developed the sparse-set coding (SSC) 
network which minimizes the number of active neurons instead of the average activity measure. Later, Olshausen et al.~\cite{olshausen2009learning} introduced an $L_1$-norm minimization criterion embedded in a highly overcomplete neural framework. Although these models offer great insight into what might be computed when receptive fields are acquired, they do not offer insight into details of the learning rules used to achieve these representations.

Early works that proposed a learning mechanism to explain the emergence of orientation
selectivity in visual cortex are those of von der Malsburg~\cite{Malsburg1973a}
and Bienenstock et al.~\cite{Bienenstock1982}.
A state-of-the-art model is that of Masquelier~\cite{masquelier2012relative}.
This model blends strong biological detail with signal processing analysis and simulation
to establish a proof-of-concept demonstration of the original 
Hubel and Wiesel~\cite{Hubel1962a} feedforward model of orientation selectivity.
A key feature of that model, relevant to the present paper,
is the use of STDP
to account for RF acquisition.
STDP is the most popular learning rule in SNNs in which the synaptic weights are adapted according to the relative pre- and postsynaptic spike times~\cite{markram2012spike,caporale2008spike}. Different variations of STDP have shown successful visual feature extraction in layer-wise training of SNNs~\cite{masquelier2007unsupervised,kheradpisheh2016bio,tavanaei2016acquisition,kheradpisheh2017stdp}.
In a similar vein,
Burbank \cite{burbank2015mirrored} has also proposed an STDP-based autoencoder.
This autoencoder uses a mirrored pair of Hebbian and anti-Hebbian STDP rules.
Its goal is to account for the emergence of symmetric, but physically separate, connections for encoding
weights ($W$) and decoding weights ($W^T$)\@.

Another component playing a key role in representing uncorrelated visual features in a bio-inspired SNN pertains to the inhibition circuits embedded within a layer. For instance, 
Savin et al.~\cite{savin2010independent} developed an independent component analysis (ICA) computation within an SNN using STDP and synaptic scaling in which independent neural activities in the representation layer were controlled by lateral inhibition. Lateral inhibition established a winner(s)-take-all (WTA) neural circuit to maintain the independence and sparsity of the neural representation layer.
More recent work~\cite{diehl2015unsupervised} has combined a layer of unsupervised STDP with an explicit layer of
non-learning inhibitory neurons.
The inhibitory neurons impose a WTA discipline.
Their representations were tested on the handwritten MNIST dataset and have been shown to be effective
for recognition of such digits.
The acquired representations tended to resemble MNIST prototypes, although their reconstructive properties
were not directly studied.
\cite{shrestha2017stable} also studied a spiking network with stochastic neurons that performs MNIST classification and acquires
MNIST prototype representations.
Their architecture is a 3-layer network where the hidden layer uses a soft
WTA to implement inhibition. 
Since there is no functional need to introduce an explicit inhibitory layer if there is no learning,
our work uses a softmax function~\cite{bishop1995neural,goodfellow2016deep} to achieve WTA inhibition.
In our work, the standard softmax is adapted to a spiking network.
Our acquired representations, when trained on the MNIST dataset, acquires representations resembling
V1-like receptive fields, in contrast to the MNIST prototypes of the research described above.

Other works related to spike-based clustering and vector quantization are the evolving
SNNs (eSNNs and deSNNs) of~\cite{wysoski2010evolving,wysoski2008fast,schliebs2013evolving,kasabov2013dynamic,soltic2010knowledge} which acquire representations via a recruitment learning paradigm~\cite{Grossberg2012adaptive}
where neurons are recruited to participate in the representation of the new pattern (based on similarity
or dissimilarity to preexisting representations).
In the deSNN framework, if a new online pattern is sufficiently similar to an already represented pattern,
the representations are merged to form a cluster.
This later work uses a number of bio-plausible mechanisms, including spiking neurons, rank-order coding
~\cite{Thorpe1998rank},
a variant of STDP, and dynamic synapses~\cite{Maass2002synapses}.

The present research proposes event-based, STDP-type rules
embedded in a single layer SNN for spatial feature coding.
Specifically, 
this paper proposes a novel STDP-based representation learning method in the spirit of~\cite{masquelier2012relative,burbank2015mirrored,zylberberg2011sparse}.
Its learning rules are local in time and space and implement an approximation to clustering-based, vector quantization~\cite{coates2012learning} using the SNN while controlling the sparseness and independence of visual codes.
Local in time means that the information to modify the synapse is recent, say within at most a couple of membrane time constant of the postsynaptic spike that triggers the STDP. By local in space, we mean that the information used to modify the synaptic weight is, in principle, available at the presynaptic terminal and the postsynaptic cell membrane.
Our derivation uses
a continuous-time formulation and takes the limit as the length
of the stimulus presentation interval tends to one time step.
This leads to STDP-type learning rules, although they differ from the classic rules found
in \cite{masquelier2012relative} and ~\cite{caporale2008spike}.
In this sense, the rules and resulting visual coding model are novel. Independence and sparsity are also maintained by an implicit inhibition and a new threshold adjustment rule implementing a WTA circuit.

\section{Background}

F\"{o}ldi\'{a}k~\cite{foldiak1989adaptive} developed a feedforward network with anti-Hebbian interconnections for visual feature extraction. The Hebbian rule in his model, shown in Eq.~\ref{oja}, is inspired from Oja's learning rule~\cite{oja1982simplified} that extracts the largest principal component from an input sequence,
\begin{equation}
\label{oja}
\Delta w_{ji} \propto (y_jx_i-w_{ji}y_j^2)
\end{equation}
\begin{equation}
y_j = \sum_i x_i w_{ji}
\label{oja2}
\end{equation}
where, $w_{ji}$ is the weight associated with the synapse connecting input (presynaptic) neuron $i$ and representation (postsynaptic) unit $j$.
$x_i$ and $y_j$ are input and linear output respectively. Over repeated trials, the term $y_jx_i$ increases the weight when the input and the output are correlated. The second term ($-w_{ji}y_j^2$) maintains the learning stability~\cite{foldiak1989adaptive}. 
With respect to binary (or spiking) units, a more appropriate assumption was made by F\"{o}ldi\'{a}k~\cite{foldiak1990forming}. He modified the previous feedforward network by incorporating non-linear threshold units in the representation layer. The units are binary neurons with a threshold of 0.5 in which $y_j\in \{0,1\}$ (Note: $y_j^2=y_j$). Thus, the Hebbian rule in Eq.~\ref{oja} is simplified to
\begin{equation}
\label{foldiakfinal}
\Delta w_{ji} \propto y_j(x_i-w_{ji}).
\end{equation} 
\begin{equation}
\label{foldiakfinal2}
y_j = 
\begin{cases}
1, & \sum_i x_i w_{ji} > 0.5 \\
0, & \mathrm{otherwise}
\end{cases}
\end{equation}

The weight change rules defined in Eqs.~\ref{oja} and~\ref{foldiakfinal} are based on the input and output correlation. Another interpretation for Eq.~\ref{foldiakfinal} can be explained in terms of vector quantization (or clustering in a WTA circuit)~\cite{hammer2002generalized,schneider2009distance} in which the weights connected to each output neuron represent particular clusters (centroids). The weight change is also affected by the output neuron activation, $y_j$. In this paper, we utilize the vector quantization concept to define an objective function. The objective function can be adapted to develop a spiking 
visual representation model equipped with a temporally local learning rule while still maintaining sparsity and 
independence. 
Our motivation is to use event-based, STDP-type learning rules. 
This requires the learning to be temporally local, specifically using spike times between pre- and 
postsynaptic neurons. 

\section{Spiking Visual Representation}
\begin{figure}
\centering
\includegraphics[scale=.6]{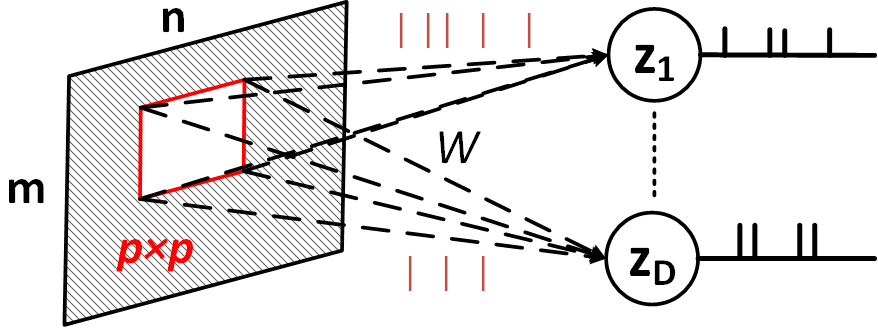}
\caption{Spiking representation network. 
$p\times p$ image patch encoded by $D$ spike trains in the representation layer. $W$ shows the synaptic weight sets corresponding to the $D$ kernels. 
}
\label{fig:network}
\end{figure}

The proposed model adopts a constrained optimization approach
to develop learning rules that are synaptically local.
The spiking representation model is a single layer SNN shown in Fig.~\ref{fig:network}.
The representation layer recodes
a $p\times p$ image patch ($p\times p$ spike trains) using $D$ spike trains generated by
neurons, $z_j$, in the representation layer. 

We derive plasticity rules that operate over a stimulus presentation interval $T$ (non-local in time) and then take the limit as 
$T$ tends to one local time step to derive event-based rules. In the case of a linear unit, $y_j$, the objective function to be minimized is shown below. It uses both the vector quantization criterion
and a regularizer that prefers small weight values.  
\begin{equation}
F(x_i,w_{ji}) = y_j(x_i- w_{ji})^2 + y_j \lambda w_{ji}^2 \ , \ \ \ \ \  y_j=\sum_i x_i w_{ji}
\label{spikeobjective}
\end{equation}
The variables $x_i,y_j,w_{ji}\geq 0$ denote: normalized input pixel intensities in the range [0, 1], the linear output activation, and the excitatory synaptic weight, respectively. The first component shows a vector quantization criterion that is scaled by the output neuron's activity, $y_j$. The $y_j$ scales the weight update rule according to the neuron's response to the input pattern ($x_i$). The second component (regularizer) is also scaled by the output neuron's activity to control the weight decay criterion (e.g. if $y_j=0$, $w_{ji}$ does not undergo learning). We assume that the input and output values can be converted to the spike counts over $T$ ms. 
The hyperparameter
$\lambda\geq 0$ controls the model's relative preference for smaller weights. 
As $\lambda \rightarrow 0$, the objective function emphasizes the vector quantization criterion. 
In contrast, as $\lambda \rightarrow \infty$ the vector quantization component is eliminated
and the minimum of the objective function is obtained when the $w_{ji}$'s $\rightarrow 0$.

In response to a stimulus presentation, a subset of spiking neurons in the representation layer is activated to code the input. To represent the stimuli by uncorrelated codes, the neurons should be activated independently and sparsely. That is, the representation layer demands a WTA neural implementation. This criterion can be achieved by a soft constraint such that
\begin{equation}
g(x) = \sum_j z_j \leq 1 \Rightarrow 1- \big (\sum_j z_j \big ) \geq 0 \;.
\label{constraint}
\end{equation}    
where, $z_j$ shows the binary state of unit $j$ after the $T$ ms presentation interval 
such that $z_j=1$ if unit $j$ fires at least once. Also, the firing status of a neuron can be controlled by its threshold, $\theta$. Therefore, this constraint can be addressed by a threshold adjustment rule.

The goal is to minimize the objective function (Eq.~\ref{spikeobjective}) while maintaining 
the constraint (Eq.~\ref{constraint}).
This can be achieved by using a Lagrangian function
\begin{equation}
L(x_i,y_j,\mathbf{z};w_{ji},\alpha) = \underbrace{y_j(x_i-  w_{ji})^2+ y_j \lambda w_{ji}^2}_{\mathrm{Objective \ Function}} - \underbrace{\alpha (1- \sum_j z_j)}_{\mathrm{Constraint}}
\label{newlogrange}
\end{equation}
where, $\alpha$ is a Lagrange multiplier.
Minimizing the first component of Eq.~\ref{newlogrange} results in a coding module that 
represents the input by a new feature vector which can cluster the data via 
the synaptic weights. 
Minimizing the second component supports the sparsity and independence of the representation to 
finally (as a special case) end with a winner-take-all network in which exactly one neuron fires upon stimulus presentation. This matter is accomplished by adapting the neuron's threshold, $\theta=-\alpha$.  
The optimum of the Lagrangian function can be obtained by gradient descent
on its derivatives 
\begin{equation}
\frac{\partial L}{\partial w_{ji}}=-2y_j (x_i - w_{ji}) + 2y_j\lambda w_{ji}
\label{deltaq1}
\end{equation} 

\begin{equation}
\frac{\partial L}{\partial \theta} = -\frac{\partial L}{\partial \alpha}= -\big (\sum_j z_j -1\big )
\label{constraint2}
\end{equation}
From gradient descent on Eq.~\ref{deltaq1} (reversing the sign on the derivative), 
we obtain
\begin{equation}
\Delta w_{ji} \propto y_j(x_i - w_{ji})-y_j\lambda w_{ji}
\label{eq:changeBasic}
\end{equation}
However, the information needed in Eq.~\ref{eq:changeBasic} is not yet temporally local.
$x_i$ denotes the rescaled pixel intensity and does not represent the input spike train. 
To re-encode a pixel intensity, $x_i$, to a spike train, $G_i$, we use uniformly 
distributed spikes (however, each spike train has a different random lag) 
with rate according to the normalized pixel intensity in the range [0, 1].
The maximum number of spikes (for a completely white pixel) over a $T=40$ ms interval is 40. 
Additionally, $y_j$ is a positive value (approximated by spike count) denoting the neuron's activation in response to a stimulus presentation and is not available at synapse, $w_{ji}$. The value $y_j$ can be reexpressed as $H_j$ representing the output spike train of neuron $j$. Spike trains $G_i$ and $H_j$ are formulated by the sum of Dirac functions as shown in Eq.~\ref{diracs}. $G_i(t)$ and $H_j(t)$ are either 0 or 1 for a given $t$.
\begin{equation}
\label{diracs}
G_i(t) = \sum_{t^{\mathrm{f}}\in S_i^{\mathrm{f}}} \delta (t-t^{\mathrm{f}}) \ \ \ \ \  , \ \ \ \ \ H_j(t) = \sum_{t^{\mathrm{f}}\in R_j^{\mathrm{f}}} \delta (t-t^{\mathrm{f}})
\end{equation} 
$S_i^{\mathrm{f}}$ and $R_j^{\mathrm{f}}$ are the sets of presynaptic and postsynaptic spike times. After coding $x_i$ and $y_j$ by spike trains $G_i$ and $H_j$ respectively, we propose a local, STDP learning rule following Eq.~\ref{eq:changeBasic}.
When, $x_i$ and $y_j$ are coded by spike trains over $T$ ms, the synaptic change in continuous time is given by
\begin{equation}
\label{continue}
\Delta w_{ji} \propto \bigg [\int_0^T H_j(t^\prime)dt^\prime \bigg ] \bigg [ \frac{1}{K}\int_0^T G_i(t^\prime)dt^\prime-w_{ji} \bigg ] -\lambda w_{ji} \int_0^T H_j(t^\prime)dt^\prime
\end{equation}
$K$ is a normalizer denoting the maximum number of presynaptic spikes over the $T$ ms interval. Over a short time period ($t\in [t^\prime , t^\prime+\gamma), \ \gamma< 1 \ \mathrm{ms}$, so that $K=1$), the weight adjustment at time $t$ is calculated by
\begin{equation}
\Delta w_{ji}(t) \ \propto \  r_j(t)\big (s_i(t)-w_{ji}(t)\big ) - \lambda w_{ji}(t)r_j(t)
\end{equation}
$r_j(t)$ shows the firing status of neuron $j$ at time $t$ ($r_j(t) \in \{0,1\}$). $s_i(t)$ specifies the presence of a presynaptic spike emitted from neuron $i$ at time interval $(t-\epsilon, t]$. In our experiments $\epsilon = 1$ ms.
The synaptic weight is changed only when a postsynaptic spike occurs ($r_j(t)=1$). 
Finally, the learning rule is formulated (upon firing of output neuron $j$) as follows
\begin{equation}
\Delta w_{ji}(t) \ \propto \  s_i(t)-w_{ji}(t)(1+\lambda)\; .
\label{eq:weightChange}
\end{equation}
Where, $w_{ji}\geq 0$. This learning rule is applied to $w_{ji}$ when postsynaptic neuron $j$ fires. The weight change is related to the presynaptic spike times received by the postsynaptic neurons. This scenario resembles spike-timing-dependent plasticity (STDP). In this STDP rule (Eq.~\ref{eq:weightChange}), the current synaptic weight affects the magnitude of the weight change. For instance, if $\lambda=0$ and $w_{ji}\in [0,1]$ (it will be proved in Eq.~\ref{stdpderive}), the smaller weights undergo larger LTP and LTD; and vice versa. It also represents a form of nearest-neighbor spike interaction~\cite{sjostrom2010spike}.

The second adaptation rule is the threshold learning rule. Eq.~\ref{constraint2} is used to implement a learning rule for adjusting the threshold, $\theta$. 
The threshold learning rule shown in Eq.~\ref{eq:thresholdrule} provides an independent and sparse feature representation. The threshold is the same for all $D$ neurons in the representation layer. 
\begin{equation}
\Delta \theta \propto \big ( \sum_{j=1}^D z_j \big ) -1
\label{eq:thresholdrule}
\end{equation}

In this section, the theory of the proposed spiking representation learning algorithm was explained. The next section will describe the SNN architecture and the learning rules derived from Eqs.~\ref{eq:weightChange} and~\ref{eq:thresholdrule}.

\section{Network Architecture and Learning}
\subsection{Neuron Model}
The network architecture is shown in Fig.~\ref{fig:network} consisting of $p^2$ and $D$ 
neurons in the input and representation layers respectively. 
Stimuli are converted to spike trains over $T$ ms for both layers. 
At a given time step,
a neuron in the representation layer is allowed to fire only if its firing criterion is met.
The firing criterion records the neuron's score in a winners-take-all competition.
The WTA score at time step t, given the entire set of incoming weights, $W$, into the representation
layer, is given by
\begin{equation}
\label{probSoftmax}
\mathit{WTAscore}_j(t; W) = \frac{\exp(\sum_i w_{ji}\zeta_i (t))}{\sum_k \exp(\sum_i w_{ki} \zeta_i(t))}
\end{equation}
\begin{equation}
\label{zeta}
\zeta_i(t) = \sum_{t^\mathrm{f}} e^{-\frac{(t-t^\mathrm{f})}{\tau}}
\end{equation}
where, $\zeta_i(t)$ is the excitatory postsynaptic potential (EPSP) generated by input 
neuron $i$ and the $t^\mathrm{f}$s are the recent spike times of unit $i$ 
during a small interval $(t - \nu, t]$, where $\nu$ is 4 ms. 
The decay time constant, $\tau$, is set to 0.5 ms.

In our network, the softmax value governs the time at which STDP occurs. 
If \textit{WTAscore} of a neuron is greater than the adaptive threshold, $\theta$, 
STDP is triggered and a spike is emitted. 
The softmax phenomologically implements mutual inhibition among the representation neurons to develop a winners-take-all circuit~\cite{tavanaei2016bio,goodfellow2016deep} in the representation layer. The neurons in the representation layer are purely excitatory and there is no explicit lateral inhibition between them other than that implicitly implemented
by the softmax. When softmax inhibition is imposed within the representation layer,
the network implements a form of competitive learning by virtue of STDP
being triggered by the firing of postsynaptic neurons.
Only neurons that ``win the competition'' are allowed to learn.

 
\subsection{Learning Rules}
The synaptic weight change shown in Eq.~\ref{eq:weightChange} defines an STDP rule where the current 
synaptic weight influences the magnitude of the change. 
STDP events are triggered upon postsynaptic firing yielding two cases corresponding to whether the presynaptic neuron fired within the $(t-\epsilon, t]$ time interval. 
Eq.~\ref{eq:stdplagrange} shows the final STDP rule derived from Eq.~\ref{eq:weightChange}. 
The weights fall in the range [0, 1] and are initialized randomly by sampling from the uniform distribution.

\begin{equation}
\label{eq:stdplagrange}
\Delta w_{ji} = 
\begin{cases}
&a \cdot \big (1-w_{ji}(1+\lambda)\big ), \ \ \mathrm{if} \ s_i=1\\
&a \cdot \big (-w_{ji}(1+\lambda)\big ), \ \ \ \ \mathrm{if} \ s_i=0
\end{cases}
\end{equation}  
\noindent
$a$ is the learning rate. If $\lambda=0$, the first and second adaptation cases increase and decrease the synaptic weight respectively (LTP and LTD). If $\lambda\rightarrow \infty$, then both cases are negative and decrease the weights down to the minimum value ($w_{ji}=0$).
Our experiments study the model's performance using different $\lambda$ values over a broad range $[0,10^{-4}, \dots ,10^4]$. Results are shown in Fig.~\ref{lambda_MNISt_W}.  

The weight adjustment, at equilibrium, reveals a probabilistic interpretation as follows
\begin{equation}
\label{stdpderive}
E[\Delta w_{ji}]=0 \leftrightarrow
\end{equation}
\begin{align*}
a\cdot P(s_i=1|r_j=1)(1-w_{ji}(1+\lambda)) - \\ a\cdot P(s_i=0|r_j=1)w_{ji}(1+\lambda) = \\
a\cdot P(s_i=1|r_j=1)(1-w_{ji}(1+\lambda))- \\ a\cdot (1-P(s_i=1|r_j=1))w_{ji}(1+\lambda) = 0 \leftrightarrow
\end{align*}
\begin{equation}
\label{stdpdrivelagrange}
(1+\lambda)w_{ji} = P(s_i=1|r_j=1)
\end{equation}
Therefore, the synaptic weight converges to the ($1+\lambda$) scaled probability of presynaptic spike occurrence given postsynaptic spike (LTP probability).
From Eq.~\ref{stdpdrivelagrange}, the weights fall in the range $(0,\frac{1}{1+\lambda})$ so that the first case refers to LTP ($\Delta w_{ji} \geq 0$) and the second one refers to LTD ($\Delta w_{ji} \leq 0$), at the equilibrium point.

To show that the STDP rule (Eq.~\ref{eq:stdplagrange})
is consistent with the learning rule in Eq.~\ref{eq:changeBasic}, we rewrite the non-local rule with learning rate, $a$, as follows
\begin{equation}
\label{consistent1}
\big (\Delta w_{ji}\big )^{\mathrm{non-local}}= a\cdot y_j\big ( x_i-w_{ji} -\lambda w_{ji} \big )
\end{equation}
As stated earlier, this rule is temporally non-local and shows the weight change over a $T$ ms interval. 
In contrast,
the STDP rule is temporally local, applying the weight change at one time step when 
the postsynaptic neuron fires.
To make Eq.~\ref{consistent1} and Eq.~\ref{eq:stdplagrange} (which is derived from Eq.~\ref{eq:weightChange}) comparable with each other, we consider a time interval with only one postsynaptic spike where $r_j=1$. Specifically, we break the $T$ ms interval into subintervals whose boundaries are determined by the event of a postsynaptic spike. It is sufficient to analyze an arbitrary subinterval. Therefore, Eq.~\ref{consistent1} at time $t$ simplifies to 
\begin{equation}
\label{consistent2}
\big (\Delta w_{ji}\big )^{\mathrm{non-local}} = a \big ( x_i-w_{ji} - \lambda w_{ji}\big )
\end{equation}
Following Eq.~\ref{stdpderive} for calculating the expected weight change using the proposed STDP rule, where $r_j=1$, we find that
\begin{equation}
\label{consistent3}
E[\Delta w_{ji}] = a\big ( P(s_i=1)-w_{ji} - \lambda w_{ji}\big )
\end{equation}
Where, $P(s_i=1)$ is the firing probability of presynaptic neuron $i$. Also, we generated the presynaptic spike trains using the normalized pixel intensities in the range [0, 1] with different random lags. 
Thus, this probability value is the same as the normalized pixel intensity, $x_i$, as firing rate. Therefore,
\begin{equation}
\label{consistent4}
E[\Delta w_{ji}] = a\big ( x_i-w_{ji} - \lambda w_{ji}\big ) = \big (\Delta w_{ji}\big )^{\mathrm{non-local}}
\end{equation}
which matches the weight change shown in Eq.~\ref{consistent2}. 
This shows that the proposed STDP rule is consistent with the non-local rule. Additionally, the STDP weight change is an unbiased estimation for the non-local (non-spike based) learning rule. Over a short time period, the proposed learning rule is also an unbiased estimation for the Hebbian rule of F\"{o}ldi\'{a}k~\cite{foldiak1990forming} (Eq.~\ref{foldiakfinal}). 
 
For the threshold adaptation, following Eq.~\ref{eq:thresholdrule}, the threshold learning rule can be written as
\begin{equation}
\Delta \theta = b\big (m_z -1 \big )
\label{eq:thresholdrule2}
\end{equation}
where, $b$ is the learning rate. $m_z$ is the number of neurons in the representation layer firing in $T$ ms. This rule adjusts the threshold such that only one neuron fires in response to a stimulus. This criterion provides a framework to extract independent features in a sparse representation.
As we used softmax-based neurons in the representation layer, the initial threshold value, $\theta^{\mathrm{init}}$, should be in the following range:

\begin{equation}
\label{initThreshold}
\frac{1}{D}<\theta^{\mathrm{init}} <<0.5
\end{equation}
\noindent
Where, $D$ is the number of neurons in the representation layer. The upper-bound of 0.5 allows more than one neuron to be active at the initial training steps to capture visual features ($\theta^{\mathrm{init}} <<0.5$). On the other hand, the initial threshold should be big enough to stop high synchronization at the beginning ($\theta^{\mathrm{init}}>\frac{1}{D}$). 
According to the minimum number of neurons we used in the experiments ($D=8, 1/D=0.125$), the initial threshold was set to 0.15. 

\section{Evaluation Metrics}
We use the following metrics to judge the quality of the representation acquired in Fig.~\ref{fig:network}.
\label{eval} 

\subsection{Reconstructed image}
We use reconstructed image to qualitatively assess the extent that the representation layer captures the information contained in the image patches.
The representation filter set, $W=\{w_1,w_2,...,w_D\}$, is a $p^2\times D$ weight matrix coding an image patch ($p^2$ input spike trains) to a vector of $D$ postsynaptic spike trains. To reconstruct the image patch from the coded spike trains, the reconstruction filter set, $W^{\mathrm{rec}}\equiv W^T$, is used to build $p^2$ spike trains. For this purpose, neurons in the input layer receive spike trains from the neurons in the representation layer via the transposed synaptic weight matrix (like an autoencoder). 


\subsection{Reconstruction loss}
To report the reconstruction loss, we use the correlation measure (Pearson correlation) and the root mean square (RMS) error between the normalized original, $\mathbf{y}_m$, and reconstructed, $\hat{\mathbf{y}}_m$, patches as shown in Eqs.~\ref{reconloss} and \ref{rms} respectively. A patch stands for $p^2$ spike rates, $\mathbf{y}$. 
\begin{equation}
\label{reconloss}
\mathit{Corr \ Recon \ Loss} = \frac{1}{M}\sum_{m=1}^M 1-Cor(\mathbf{y}_m,\hat{\mathbf{y}}_m)
\end{equation} 
\begin{equation}
\label{rms}
\mathit{RMS} =  \frac{1}{M} \sum_{m=1}^M \sqrt{\frac{1}{p^2}\sum_{i=1}^{p^2} (y_{i,m}-\hat{y}_{i,m})^2} \; .
\end{equation}
$M$ is the number of patches extracted from the image.

\subsection{Sparsity}
To calculate the sparsity,
we use average activity and breadth tuning measures. The average activity specifies the density of spikes released from neurons in the representation layer over $T$ time steps given in Eq.~\ref{sparse1}.
\begin{equation}
\label{sparse1}
\mathit{Sparsity} = \frac{1}{D\cdot T} \sum_j\sum_t r_j(t) \; .
\end{equation}  
The breadth tuning measure introduced by Rolls and Tovee~\cite{rolls1995sparseness} specifies the density of neural layer activity (Eq.~\ref{sparse2}) calculated by the ratio of mean, $\mu$, and standard deviation, $\sigma$, of spike frequencies in the representation layer upon presenting a stimulus. The breadth tuning measures the neural selectivity such that the sparse code distribution concentrates near zero with a heavy tail~\cite{Foldiak2008sparse}. For a neural layer where most of the neurons fire, the activity distribution is more uniformly spread and $\mathit{Breadth \ Tuning}$ is greater than 0.5. In contrast, in a sparse code where most of the neurons do not fire, the distribution is peaked at zero and $\mathit{Breadth \ Tuning}$ is less than 0.5. 
\begin{equation}
\label{sparse2}
\mathit{Breadth \ Tuning} = \frac{1}{C^2+1} , \ \ C = \frac{\sigma}{\mu}
\end{equation} 

\section{Experiments and Results}
We ran two experiments using the MNIST~\cite{lecun1998mnist} and the natural image~\cite{olshausen1996emergence} datasets to evaluate the proposed local representation learning rules embedded in the single-layer SNN. For both datasets, the intensities of the gray-scale images were normalized to fall in the range of [0, 1] yielding possible spike rates to generate uniformly distributed spike trains for the input layer over $T=40$ ms. The learning rates for STDP learning, $a$, (Eq.~\ref{eq:stdplagrange}) and for threshold adjustment, $b$, (Eq.~\ref{eq:thresholdrule2}) were set to 0.0005 and 0.0001, respectively\footnote{The maximum number of postsynaptic spikes is 40 and the maximum number of patches sampled from an MNIST digit is 25. Our simple strategy for setting the learning rates is: $a,b <\frac{1}{25\times 40}=0.001$.}.
We ran a number of experiments with different learning rates and found that changing $a$ and $b$ in the range [0.00005, 0.001] did not change the model's performance significantly. Additionally, as the threshold adjustment rule is not modulated by the current threshold value, we chose a smaller learning rate (0.0001) for it to avoid possible threshold instability.

\subsection{Experiment 1: MNIST dataset}
\begin{figure}
\centering
\subfloat[]{
\includegraphics[scale=.32]{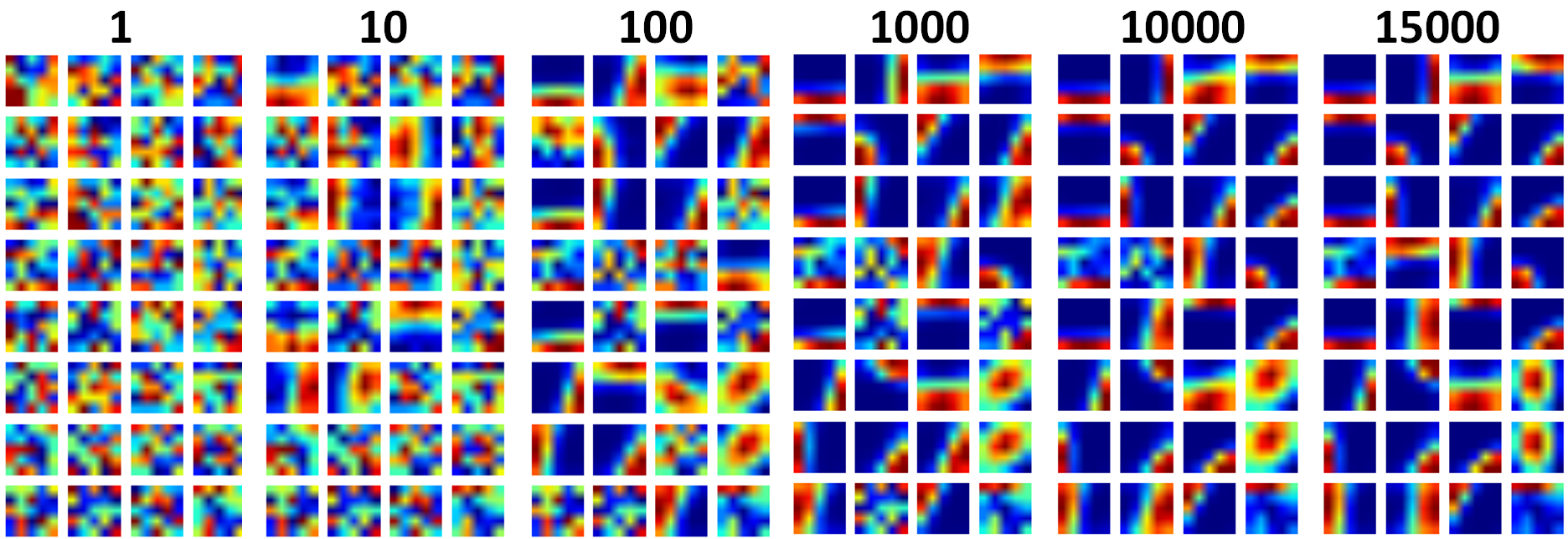}
\label{fig:filters32}
}
\quad
\subfloat[]{
\includegraphics[scale=.25]{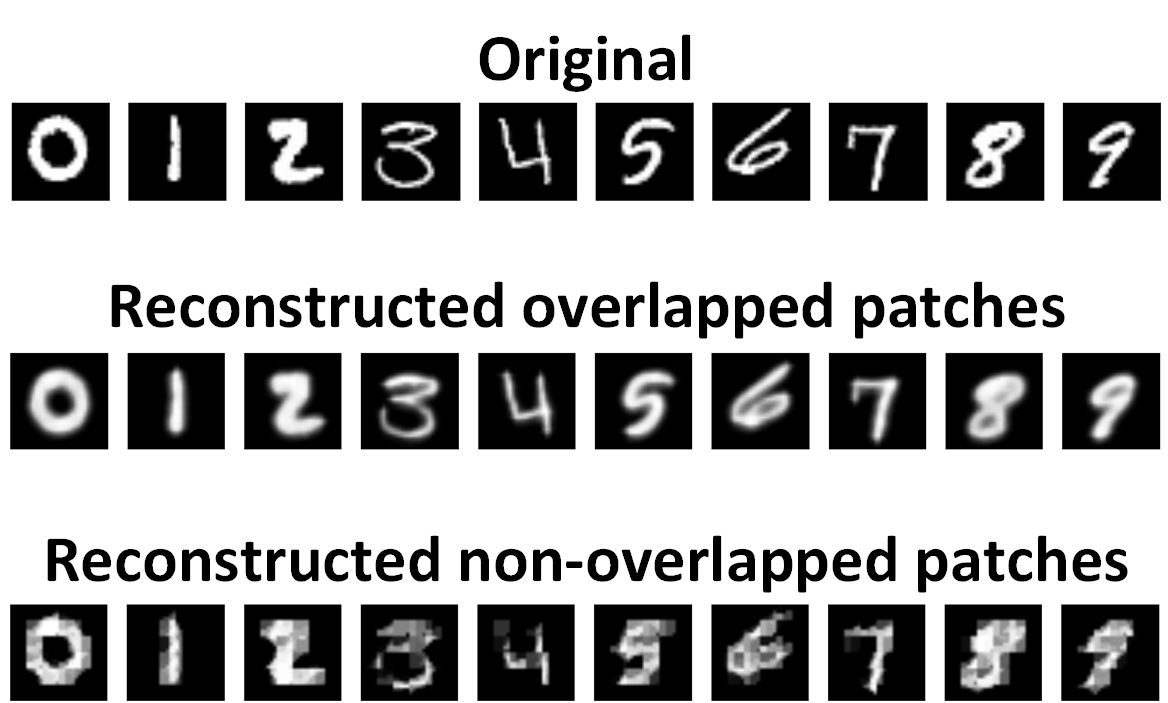}
\label{fig:recon_32_mnist}
}
\quad
\subfloat[]{
\includegraphics[scale=.35]{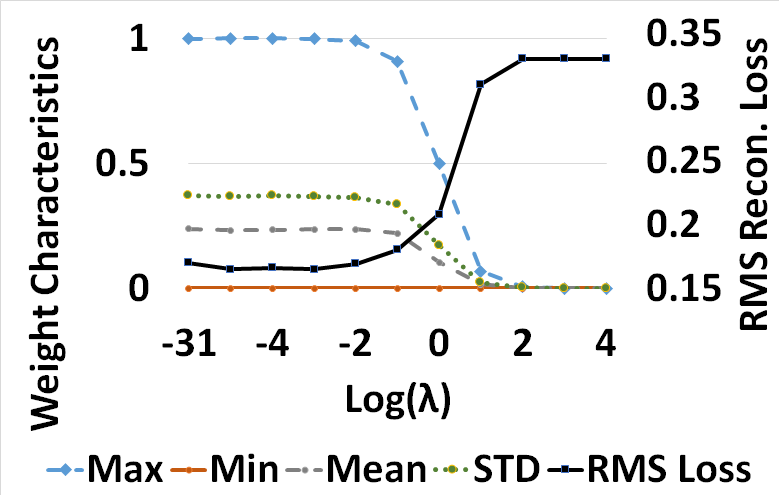}
\label{lambda_MNISt_W}
}
\caption{(a): $D=32$ trained filters after 1, 10, ..., 15000 iterations. The red-blue spectrum denotes the maximum-minimum synaptic weights. (b): Reconstructed images based on overlapped and non-overlapped $5\times 5$ patches. The overlapped patches are selected by $5\times 5$ windows sliding over the image with a stride of 1. The non-overlapped patches slide over the image with a stride of 5. (c): RMS reconstruction loss and synaptic weight ranges for the SNN with $D=32$ filters versus $\log_{10}$ regularizer hyperparameter, $\lambda$. $\lambda=0$ is approximated by $10^{-31}$.}

\end{figure}

Experiments were run using $5\times 5$ patches sampled from $28\times 28$ MNIST digits. 
We used a random subset of the MNIST digits divided into 15,000 training and 1000 testing images for learning and evaluating the model. 
The SNN architecture consists of 25 ($5\times 5$ image patch) neurons in the input layer and $D=\{2^i, \ i=3\dots 7\}$ neurons in the representation layer. These variations of the network architecture (different $D$ values) determine under-complete to over-complete representations. Trained filters, after 1 through 15,000 iterations, for the network with 32 neurons in the representation layer are shown in Fig.~\ref{fig:filters32}. After 1000 training iterations, the kernels start becoming selective to specific visual patterns (orientations). 
The filters shown in this image tend to be orientation selective and extract different visual features.

Fig.~\ref{lambda_MNISt_W} shows the RMS reconstruction loss and other statistical characteristics (max, min, mean) of the trained weights versus the log regularizer hyperparameter ($\log_{10}\lambda$).
For $\lambda\leq 0.1$, the RMS loss values reach a near optimal uniformly low plateau\footnote{The average RMS reconstruction loss values for the SNNs with $\lambda \leq 0.1$ was reported $0.167\pm 0.001$ (Ninety-five percent confidence intervals of the RMS loss values (standard error of the mean; $n=5$) were calculated).}. For this reason, $\lambda$ was set to zero for further experiments.
Additionally, Fig.~\ref{lambda_MNISt_W} shows that the maximum and minimum synaptic weights after training are $1/(1+\lambda)$ and 0 
respectively as predicted by Eq.~\ref{stdpdrivelagrange}. 

\begin{figure}
\centering
\includegraphics[scale=.3]{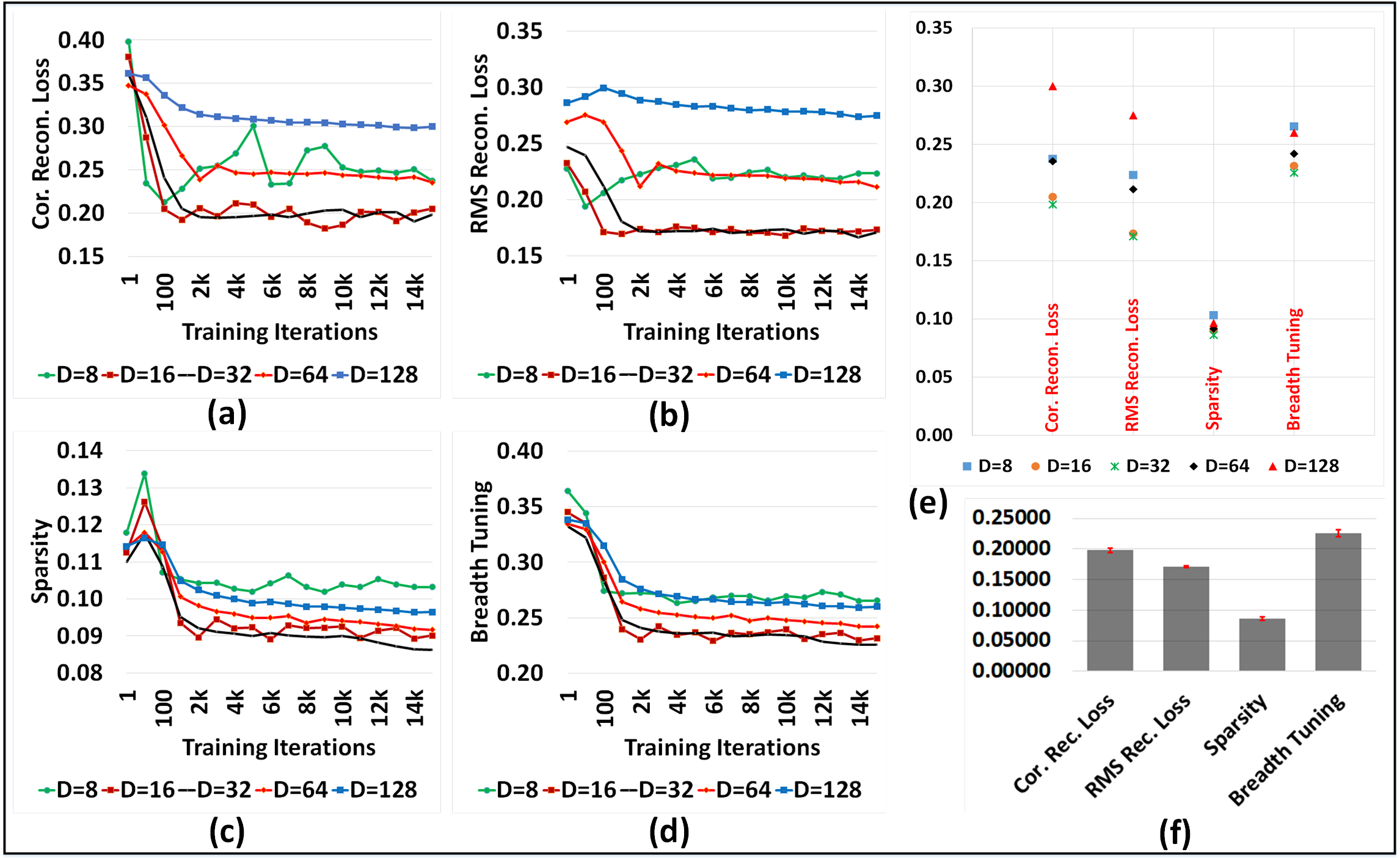}
\caption{\textbf{(a)-(d)}: Model performance trends on MNIST after 1 through 15000 training iterations in terms of (a): correlation-based reconstruction loss, (b): RMS reconstruction loss, (c): average sparsity, and (d): breadth tuning.\textbf{(e)}: The model's performance after training. \textbf{(f)}: The evaluation measures for the trained visual representation model with $D=32$ kernels. Error bars show standard error of the mean.}
\label{fig:processMNIST}
\end{figure}

The three performance measures from the Section 5 were used to assess the model.
These were
the reconstructed images, the reconstruction loss, and the sparsity.
The reconstructed images of randomly selected digits 0 through 9, acquired by the SNN with $D=32$ neurons in the representation layer, are shown in Fig.~\ref{fig:recon_32_mnist}. The reconstructed maps show high quality images comparable with the original images. 
The reconstruction loss measures for the SNNs with $D=8$ through $128$ filters appear in Figs.~\ref{fig:processMNIST}a and \ref{fig:processMNIST}b. 
The SNNs with $D=16$ and $32$ show the lowest reconstruction loss after training.  
The sparsity measures reported by the average sparsity and the breadth tuning are shown in Fig.~\ref{fig:processMNIST}c and~\ref{fig:processMNIST}d. The sparsity measures also show better performance for the networks with $D=\{16,32,64\}$ filters. The average sparsity value of 0.09 shows that only 9\% of the neurons were active on each trial. 
The breadth tuning value of 0.23 indicates the sparse stimulus representation.
Figs.~\ref{fig:processMNIST}e and~\ref{fig:processMNIST}f depict the summary of the model's performance after training for $D=\{8,\dots ,128\}$ and $D=32$ kernels, respectively.     

\subsection{Experiment 2: Natural images}
\begin{figure}
\centering
\subfloat[]
{
\includegraphics[scale=.25]{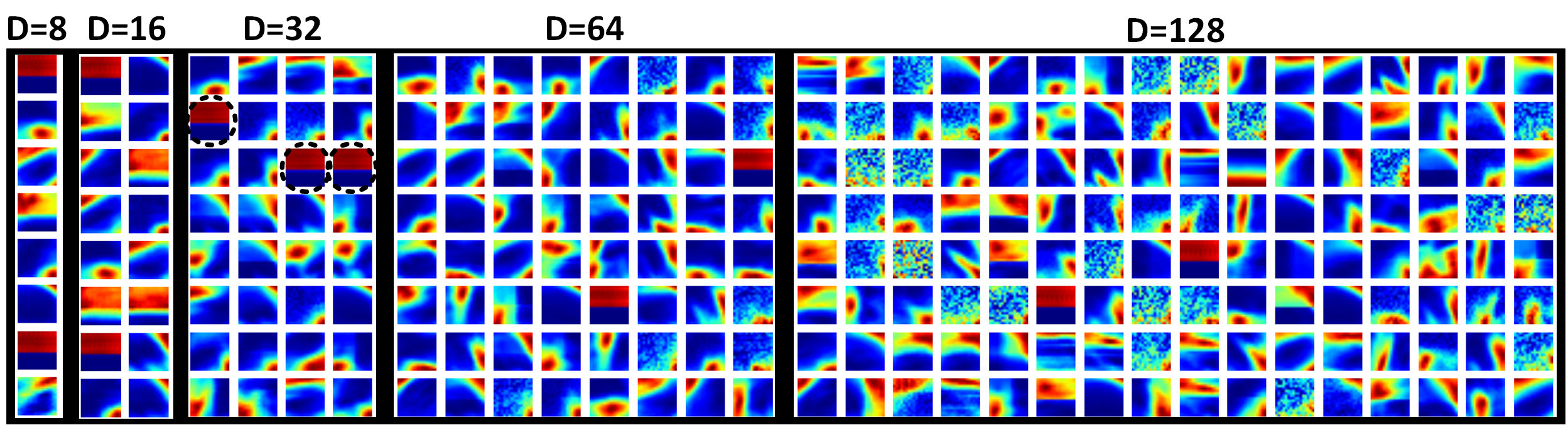}
\label{fig:nature1}
}
\quad
\subfloat[]
{
\includegraphics[scale=.55]{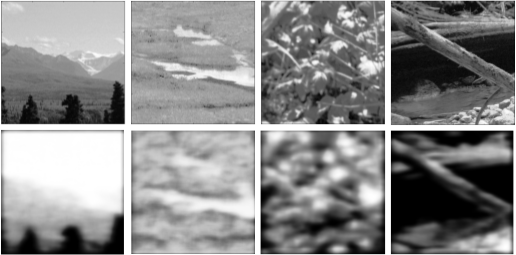}
\label{fig:nature3}
}
\quad
\subfloat[]
{
\includegraphics[scale=.35]{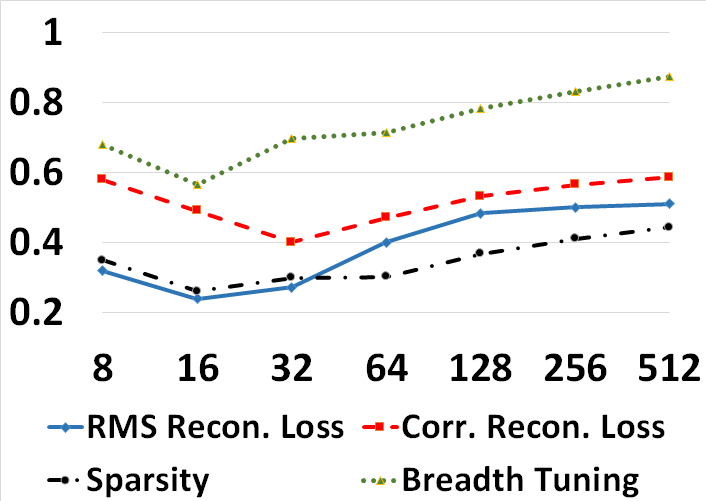}
\label{fig:nature2}
}
\label{fig:nature}
\caption{Model's performance on the natural image patches. (a):  Representation filters after training the SNNs with $D=\{8,16,32,64,128\}$ spiking neurons in the representation layer. (b): Original (first row) and reconstructed (second row) image sections ($D=32$). (c): Reconstruction loss and sparsity measures of the models with 8 through 512 filters.}
\end{figure}

This experiment evaluates representations acquired from $16\times 16$ natural image patches \cite{olshausen1996emergence}. 
Fig.~\ref{fig:nature1} shows the trained representation filters for the SNNs with 8, 16, 32, 64, and 128 neurons in the representation layer. For instance, where $D=32$, except for the filters marked with dotted circles, the other filters have low correlation with each other. 
For visual assessments, Fig.~\ref{fig:nature3} shows four natural images and their reconstructed maps.
Performance of the proposed model in terms of the reconstruction loss and sparsity measures on natural images is shown in Fig.~\ref{fig:nature2}. The lowest reconstruction loss belongs to the networks with $\{16,32,64\}$ neurons in the representation layer. The small number of neurons ($D=8$) is not able to capture the visual codes. On the other hand, using too many neurons ($D>128$) increases reconstruction loss because a number of neurons cannot be involved in the learning process due to the WTA constraint.   

\subsection{Comparisons}
\begin{figure}
\centering
\includegraphics[scale=.26]{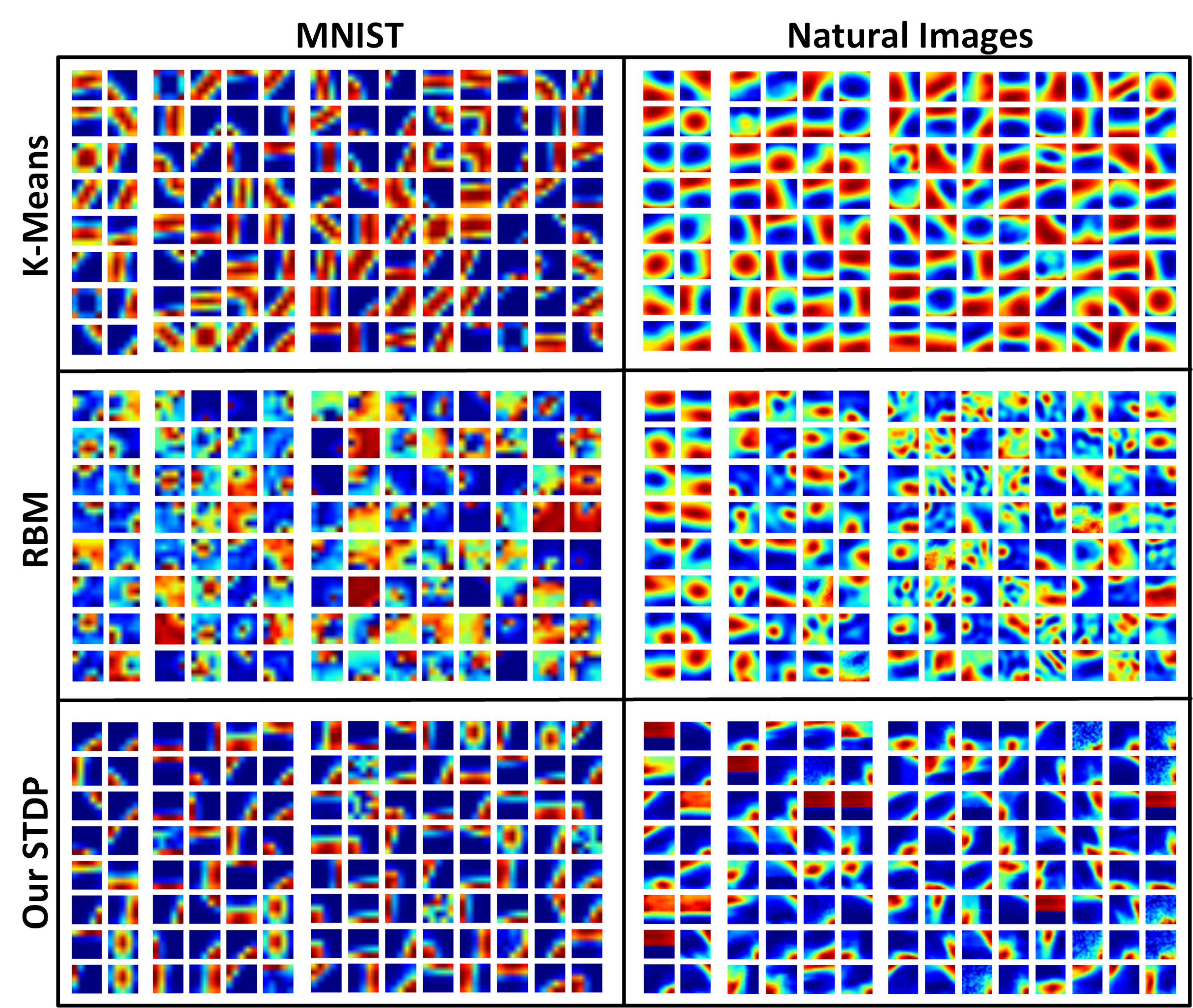}
\caption{$D=16,32,$ and $64$ representation filters trained on the MNIST and natural images datasets using K-means, RBM, and our STDP.}
\label{fig:compare}
\end{figure}

The proposed spiking representation learning method shows better performance than the traditional K-means clustering~\cite{bishop2006pattern} and the restricted Boltzman machine (RBM)~\cite{hinton2010practical,le2008representational} while introducing local learning in time and space. We implemented these two methods, as traditional quantization-like representation learning examples, using the same training/testing images.
The K-means and RBM approaches were applied to the normalized pixel intensities of image patches (not spike trains). Thus, these methods are not temporally local.
Table~\ref{tab:compare} shows this comparison in terms of reconstruction loss (correlation and RMS). Our model outperforms the RBM and K-means methods except for the two cases (natural images) in which the RBM shows slightly better performance. Fig.~\ref{fig:compare} shows the trained filters obtained by K-means, RBMs, and our model based on the MNIST and natural image patches. K-means, similar to our model, detects different visual orientations for the MNIST and natural image patches, but the filters are highly correlated. The RBM did not perform well for the MNIST dataset but it successfully learned representative visual filters for the natural image patches where $D=64$. These trained filters (Fig.~\ref{fig:compare}) confirm the reconstruction loss variations reported in Table~\ref{tab:compare}.    

\begin{table}[]
\centering
\footnotesize
\caption{Reconstruction loss (correlation and RMS) obtained by K-means and RBM in comparison with our method.}
\label{tab:compare}
\begin{tabular}{|l|l|l|l|l|l|l|l|l|l|l|l|l|}
\hline
\multicolumn{1}{|c|}{Rec. Loss}  & \multicolumn{3}{c|}{MNIST Corr.}                                                                      & \multicolumn{3}{c|}{MNIST RMS}                                                                        & \multicolumn{3}{c|}{Natural Corr.}                                                                     & \multicolumn{3}{c|}{Natural RMS}                                                                       \\ \hline
\multicolumn{1}{|c|}{\textit{D}} & \multicolumn{1}{c|}{\textit{16}} & \multicolumn{1}{c|}{\textit{32}} & \multicolumn{1}{c|}{\textit{64}} & \multicolumn{1}{c|}{\textit{16}} & \multicolumn{1}{c|}{\textit{32}} & \multicolumn{1}{c|}{\textit{64}} & \multicolumn{1}{c|}{\textit{16}} & \multicolumn{1}{c|}{\textit{32}} & \multicolumn{1}{c|}{\textit{64}} & \multicolumn{1}{c|}{\textit{16}} & \multicolumn{1}{c|}{\textit{32}} & \multicolumn{1}{c|}{\textit{64}} \\ \hline
K-means                          & 0.22                             & 0.23                             & 0.26                             & 0.18                             & 0.21                             & 0.26                             & \textbf{0.45}                             & 0.52                             & 0.57                             & 0.31                             & 0.36                             & 0.40                              \\ \hline
RBM                              & 0.49                             & 0.49                             & 0.40                              & 0.27                             & 0.27                             & 0.26                             & 0.92                             & 0.41                             & \textbf{0.44}                    & 0.47                             & \textbf{0.27}                    & \textbf{0.26}                    \\ \hline
\textbf{Our STDP}                & \textbf{0.20}                     & \textbf{0.20}                     & \textbf{0.24}                    & \textbf{0.17}                    & \textbf{0.17}                    & \textbf{0.21}                    & 0.49                    & \textbf{0.40}                     & 0.47                             & \textbf{0.24}                    & \textbf{0.27}                    & 0.40                              \\ \hline
\end{tabular}
\end{table}

Table~\ref{tab:compare2} compares our results with the only (to the best of our knowledge) spike-based representation learning models. The correlation-based reconstruction loss on MNIST and natural images (0.2 and 0.4) shows improvement over the existing spiking autoencoder using mirrored STDP (0.2 and 0.65) proposed by Burbank~\cite{burbank2015mirrored}. The sparse representation introduced by King et al~\cite{king2013inhibitory}, which is a modified version of the SAILnet algorithm~\cite{zylberberg2011sparse}, reported an RMS reconstruction loss around 0.74 that is calculated based on the spike rates normalized to unit standard deviation (let's say zRMS). Our model compared favorably with their model with zRMS=0.67. However, our model did not scale well to a larger number of neurons when $D\geq128$ in the representation layer. The problem appears to stem from the threshold adjustment rule (Eq.~\ref{eq:thresholdrule2}). If we change the rule to $\Delta \theta = b\big (m_z -q \big )$, where $q$ is a proportion of $D$, the representation layer would be more active and a large number of filters can be trained to reduce the reconstruction loss.      

\begin{table}[]
\centering
\footnotesize
\caption{The reconstruction loss values reported by Burbank~\cite{burbank2015mirrored} in terms of correlation loss and King et al.~\cite{king2013inhibitory} in terms of zRMS in comparison with our results.}
\label{tab:compare2}
\begin{tabular}{|l|l|c|l|}
\hline
\multicolumn{1}{|c|}{\textbf{Dataset}} & \multicolumn{1}{c|}{\textbf{Burbank~\cite{burbank2015mirrored}}} & \textbf{King et al.~\cite{king2013inhibitory}}             & \multicolumn{1}{c|}{\textbf{Our Model}} \\ \hline
Natural images                                 & Corr: 0.65                            & \multicolumn{1}{l|}{zRMS: 0.74} & Corr: \textbf{0.4}, zRMS: \textbf{0.67}                  \\ \hline
MNIST                                  & Corr: \textbf{0.2}                             & -                                & Corr: \textbf{0.2}                               \\ \hline
\end{tabular}
\end{table}

\section{Discussion and Conclusion}
This paper derived a novel STDP-based representation learning method to be embedded in an SNN and evaluated the acquired representations in two experiments to establish the method's initial viability.
The derived rules were extremely simple, yet the evaluated reconstruction loss was extremely low. The simplicity of the rules (resulting from the constraint of temporal locality) makes them attractive for hardware implementation.

The learning rules were derived by constrained optimization incorporating a vector quantization-like objective function with regularization and a sparsity constraint. The learning rules included spatio-temporally local STDP-type weight adaptation and a threshold adjustment rule. The STDP rule at equilibrium showed a probabilistic interpretation of the synaptic weights scaled by the regularizer hyperparameter. In addition to the threshold adaptation rule, the WTA-thresholded neurons in the representation layer implemented inhibition (by a novel temporal, spiking softmax function) to represent sparse and independent visual features. Softmax is a standard way to implement a winners-take-all (WTA) circuit and to implement mutual inhibition without using explicit inhibitory neurons in the representation layer~\cite[p. 181]{goodfellow2016deep},~\cite[p. 238]{bishop1995neural}.

The experimental results showed high performance of the proposed model in comparison with spiking and non-spiking approaches. Our model almost outperformed the traditional K-Means and RBM models in representation learning and training of the orientation selective kernels. Also, our method showed better performance (in terms of reconstruction loss) than the state-of-the-art spiking representation learning approaches used by~\cite{burbank2015mirrored} (spiking autoencoder) and~\cite{zylberberg2011sparse,king2013inhibitory} (sparse representation). 

To obtain the spatio-temporally local learning rules embedded in the SNN, we started from a non-spiking quantization criterion inspired from~\cite{foldiak1990forming}. Then, we developed novel rules to implement an STDP based representation learning and a threshold adjustment rule for spiking platforms. The spike-based platform and spatio-temporally local learning rules lead the main difference between our study and well-known, traditional representation learning methods introduced in the literature. Existing spiking representation learning methods in the literature suffer from limitations such as violating Dale's law~\cite{zylberberg2011sparse}, synapses that can change sign~\cite{zylberberg2011sparse,king2013inhibitory}, low performance in terms of reconstruction loss~\cite{burbank2015mirrored}, and non-spiking input signals~\cite{zylberberg2011sparse,king2013inhibitory}. In this study we proposed an STDP learning rule which updates the synaptic weights falling within the range [0, 1]. The SNN architecture consists of excitatory neurons and an implicit inhibition occurring in the representation layer. The implicit inhibition is analogous to a separate inhibitory layer balancing neural activities in the representation neural layer where Dale's law is maintained. Furthermore, the proposed SNN implements spiking neurons in both the input and representation layers and the neurons only communicate through temporal spike trains.

To the best of our knowledge, our approach is the only high performance (in terms of reconstruction loss) representation learning model implemented on SNNs. There are several studies in the literature developing SNNs equipped with bio-inspired STDP for unsupervised feature extraction through single or multi-layer spike-based architectures. The most recent works of~\cite{diehl2015unsupervised,shrestha2017stable,kheradpisheh2017stdp}, and~\cite{tavanaei2017multi} have utilized these features to classify MNIST digits. Although these networks introduce novel spiking network architectures for feature representation, they do not offer a pure representation learning approach with low reconstruction loss. 

Although our proposed spiking representation learning was successful for reconstruction, there is a limitation that the spike rate of the presynaptic neurons is higher than biological spiking neurons. Our future work seeks to reduce this spike rate to be more biologically plausible. Using more presynaptic neurons presenting mutual exclusive intensity bands would be a starting point. Additionally, it is a matter of future work to determine how well the acquired representations from our STDP algorithm perform in a pattern recognition context. It can also be tested in future work whether our acquired representations are stackable to afford the ability for multi-layer, STDP-based learning.

\bibliographystyle{elsarticle-num}
\bibliography{references}

\end{document}